\documentclass[sigconf]{acmart}

\usepackage{multirow}
\AtBeginDocument{%
  }

\copyrightyear{2025}
\acmYear{2025}
\setcopyright{acmlicensed}
\acmConference[KDD '25] {Proceedings of the 31st ACM SIGKDD Conference on Knowledge Discovery and Data Mining V.2}{August 3--7, 2025}{Toronto, ON, Canada.}
\acmBooktitle{Proceedings of the 31st ACM SIGKDD Conference on Knowledge Discovery and Data Mining V.2 (KDD '25), August 3--7, 2025, Toronto, ON, Canada}
\acmISBN{979-8-4007-1454-2/25/08}
\acmDOI{10.1145/3711896.3737203}

\begin{document}
\newcommand{\pheadNoSpace}[1] {\noindent\textbf{#1.}} 
\newcommand{\pheadWithSpace}[1] {\vspace{1.25mm}\noindent\textbf{#1.}} 
\title{\textit{COEF-VQ}: Cost-Efficient Video Quality Understanding through a Cascaded Multimodal LLM Framework}

\author{Xin Dong}
\email{xindong@bytedance.com}
\affiliation{%
  \institution{ByteDance Inc.}
  \city{San Jose}
  \state{CA}
  \country{United States}}

\author{Sen Jia}
\email{jiasen.bit@bytedance.com}
\affiliation{%
  \institution{ByteDance Inc.}
  \city{Beijing}
  \state{}
  \country{China}}

\author{Ming Rui Wang}
\email{mingrui.wang2@bytedance.com}
\affiliation{%
  \institution{ByteDance Inc.}
  \city{Singapore}
  \state{}
  \country{Singapore}}

\author{Yan Li}
\email{liyan.1994@bytedance.com}
\affiliation{%
  \institution{ByteDance Inc.}
  \city{Beijing}
  \state{}
  \country{China}}

\author{Zhenheng Yang}
\email{yangzhenheng@bytedance.com}
\affiliation{%
  \institution{ByteDance Inc.}
  \city{San Jose}
  \state{CA}
  \country{United States}}

\author{Bingfeng Deng}
\email{dengbingfeng@bytedance.com}
\affiliation{%
  \institution{ByteDance Inc.}
  \city{Singapore}
  \state{}
  \country{Singapore}}

\author{Hongyu Xiong}
\authornote{Author corresponded for this research.}
\email{hongyu.xiong@bytedance.com}
\affiliation{%
  \institution{ByteDance Inc.}
  \city{San Jose}
  \state{CA}
  \country{United States}}

\renewcommand{\shortauthors}{Xin Dong et al.}

\begin{abstract}
Recently, with the emergence of recent Multimodal Large Language Model (MLLM) technology, it has become possible to exploit its video understanding capability on different classification tasks. 
In practice, we face the difficulty of huge requirements for GPU resource if we need to deploy MLLMs online. 
In this paper, we propose \textit{COEF-VQ}, a novel cascaded MLLM framework designed to enhance video quality understanding on the short-video platform while optimizing computational efficiency. Our approach integrates an entropy-based pre-filtering stage, where a lightweight model assesses uncertainty and selectively filters cases before passing them to the more computationally intensive MLLM for final evaluation. By prioritizing high-uncertainty samples for deeper analysis, our framework significantly reduces GPU usage while maintaining the strong classification performance of a full MLLM deployment.
To demonstrate the effectiveness of \textit{COEF-VQ}, we deploy this new framework onto the video management platform (VMP) at the short-video platform, and perform a series of detailed experiments on two in-house tasks related to video quality understanding. 
We show that \textit{COEF-VQ} leads to substantial performance gains from the offline evaluation in these two tasks and effectively enhances platform safety with limit resource consumption, significantly reducing inappropriate content video view rate by 9.9\% in a online A/B test without affecting engagement. Post-launch monitoring confirmed sustained improvements, validating its real-world impact.
\end{abstract}



\maketitle

\begin{figure*}[tbhp]
    \centering
	\includegraphics[width=0.9 \linewidth]{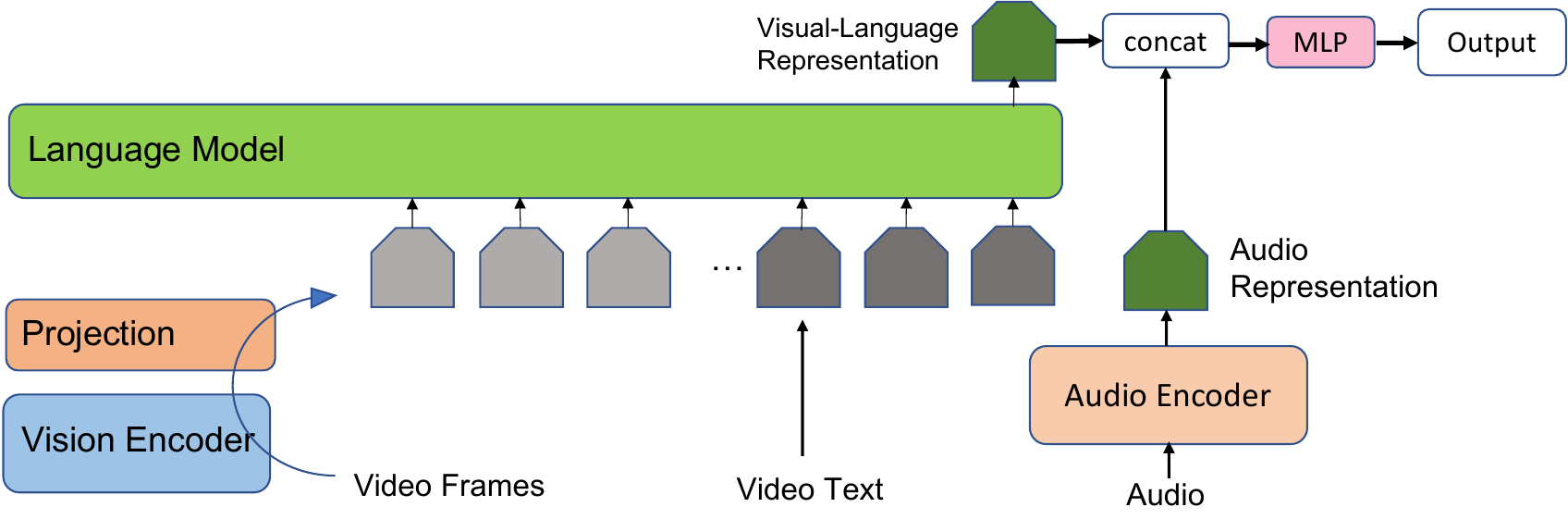}
	\caption{Illustration of our proposed unified Multimodal LLM architecture: video frames and text tokens are processed and early-fused through a VL foundational model, and audio signal is processed through an Audio encoder and then late-fused with the hidden representations from VL model.}
	\label{fig:mllm}
\end{figure*}

\section{Introduction}
In current video understanding models, besides learning visual information from video frames, additional data signals such as texts (captions, subtitles, or on-screen texts) and audio (dialogues, background music, or sound effects) are commonly used to capture a more comprehensive understanding of the video \cite{conneau2019unsupervised,he2016deep,radford2023robust}. 
Industry practices often deploy a multi-tower late-fusion architecture \cite{radford2021learning,zhai2023sigmoid}, where each tower independently processes a single modality and then all modality outputs are fused to obtain a final prediction. 
For instance, one tower might process visual frames, another audio signals, and a third tower textual data. The separate feature representations generated by each tower are typically combined, or fused, in later layers of the model to produce a more robust and holistic classification output. 
This approach allows the model to leverage the complementary nature of different modalities to improve performance.

With the rapid advancement of large language models (LLMs) and multimodal large language models (MLLMs) \cite{achiam2023gpt,team2023gemini}, these models have demonstrated remarkable capabilities in a broad range of applications, including question answering on image or video content \cite{li2024llava, wang2024qwen2}, commonsense reasoning \cite{li2024llava, wang2024qwen2}, and contextual comprehension \cite{li2021align, wang2024qwen2}. 
They excel at recognizing and interpreting complex relationships across modalities, improving the model’s general capabilities in handling various multimodal tasks. 
However, the effectiveness of these models in highly specialized or sensitive domains is still under investigation.
For example, specific tasks such as evaluating whether a video has undergone professional editing, or detecting whether a video violates platform's trust and safety (TnS) policy, require an in-depth understanding of nuanced and domain-specific characteristics. 
These are often hard-to-detect signals that need a fine-grained understanding beyond general visual and textual comprehension. 

Therefore, in this paper we propose \textit{COEF-VQ}, which is composed of two key novelties:

\( 1) \) we first design a unified Multimodal LLM framework integrating video frames, texts, and audio information, which goes beyond the existing multi-tower baseline deployed on our platform \cite{conneau2019unsupervised,he2016deep,liu2021swin,radford2023robust} and typical industry practices \cite{zhai2023sigmoid}. 
This enhanced model architecture allows for more sophisticated feature fusion and deeper cross-modal interactions, boosting the video classification performance. 
By doing so, it alleviates the limitations of traditional multi-tower structures, making the model more sensitive and accurate in discerning domain-specific tasks.

However, despite the powerful capabilities of MLLM, the large number of parameters in MLLM, which enables them to perform robust multimodal reasoning, requires substantial GPU memory and processing speed. 
This demand, when scaled for online deployment on short-video platforms, would lead to considerable GPU resource consumption, affecting both deployment costs and latency in real-time, online video service.

\( 2) \) 
we develop a cascaded serving framework for online deployment, designed to optimize resource utilization while maintaining strong performance. In the first stage, a lightweight model (typically the baseline) generates initial predictions at the time of video posting. To enhance filtering efficiency, we introduce an entropy-based criterion, where videos with low uncertainty are filtered out, significantly reducing the volume passed to the next stage. This ensures that only uncertain cases undergo further evaluation.  
In the second stage, an MLLM processes the remaining videos with higher accuracy, effectively reducing false positives. This entropy-driven pre-filtering not only minimizes unnecessary computational overhead but also ensures that the MLLM focuses on the most ambiguous or critical cases, improving classification reliability. By prioritizing MLLM processing for high-uncertainty inputs, our approach achieves a scalable and cost-effective deployment of MLLMs in real-time environments. This framework is particularly well-suited for high-stakes applications, such as video quality understanding on short-video platforms, where balancing resource efficiency with classification accuracy is crucial.

By deploying \textit{COEF-VQ} framework on internal video management plaform (VMP) from the short-video platform, our empirical study on two in-house tasks demonstrates the cost-efficiency of incorporating MLLM into a cascade framework and the superiority of our new cascade method on domain-specific classification tasks. We also conduct a series of experiments on Inappropriate Content Detection (ICD) for online deployment, a key task in maintaining platform safety. Our model deployment reduces GPU usage to only 5\% of a full-traffic MLLM deployment during online A/B testing. Over a 10-day online experiment with 10\% traffic allocation per group, our framework significantly reduced the inappropriate content video view rate (\textbf{ic\_vvr}) by 9.9\% without impacting core engagement metrics. The post-launch evaluation, spanning more than a month of global monitoring, further confirmed the effectiveness of COEF-VQ, demonstrating sustained reductions in \textbf{ic\_vvr} and validating its real-world impact.

\section{Methodology}

\subsection{Multimodal LLM Architecture}
The Multimodal LLM architecture, as illustrated in \autoref{fig:mllm}, inherits the design of vision-language (VL) foundation model combined with audio encoder. 
Several key components are integrated to process multimodal inputs, including a vision encoder for video frames, a VL modality alignment projector, a language model for both vision and text tokens, and an audio encoder for the video’s audio stream.

\pheadWithSpace{Vision Encoder for Video Frames}
   For each frame in a video, the input frame \( X_f \) is passed through a vision encoder to extract its visual features. The encoder function $g(\cdot; \theta)$  transforms the frame into a visual feature representation:
   \begin{equation}
   Z_f = g(X_f; \theta)
   \end{equation}
   where \( Z_f \) captures the essential visual information from each frame, allowing subsequent layers to process this information effectively.

\pheadNoSpace{Modality Alignment Projector}
   A projector module is used to align visual and textual modalities, creating a unified representation space. This alignment enables the model to handle multiple modalities consistently, ensuring that video frames and text features are comparably represented and effectively combined. In practice, a 2-layer MLP $p(\cdot; \theta)$ yields a sequence of visual token representations:
   \begin{equation}
   H_f = p(Z_f; \theta)
   \end{equation}

\pheadNoSpace{Language Model for Vision and Text Tokens}
   The model processes a sequence of tokens with length \( L \), consisting of both patch tokens \( H_f \) (derived from video frames) and text tokens \( X_t \) (representing video text, such as title or stickers). The large language model (LLM) $lm(\cdot; \theta)$ integrates these multimodal tokens, and the hidden representation of the final token in this sequence, denoted as \( H \), encodes combined information from video frames and associated video text. 
   \begin{equation}
   H = lm((H_f,X_t); \theta)
   \end{equation}
\pheadNoSpace{Audio Encoder for Video's Audio}
   To incorporate audio information, an audio encoder $a(\cdot; \theta)$ processes the raw audio input from the video, denoted as \( X_a \). Average pooling is invoked for yield the hidden representation of the audio signal. 
   \begin{equation}
   H_a = avg\_pooling(a(X_a; \theta))
   \end{equation}
   This audio encoder extracts audio features relevant to the classification task, allowing the model to incorporate sound cues, which are particularly useful when videos include narration or other informative audio elements.

\pheadNoSpace{Fusion and Classification Layer}
   The output of the audio encoder \( H_a \) is then concatenated with the hidden representation \( H \) from the language model. This combined embedding, which merges visual, textual, and audio information, is denoted as \(H_{cl} = [H; H_a]\)
   The classification module $f_\mathrm{cl}(\cdot; \theta_\mathrm{cl})$ is applied  on top of the combined embedding \(H_{cl}\). This module consists of a linear function mapping $H_{cl} \in \mathbb{R}^{d}$ into $\mathbb{R}^{|\mathcal{Y}|}$ and a softmax function, where $\mathcal{Y}$ is the set of target classes. 

Our MLLM combines video frames, text, and audio into a single, cohesive representation that enables rich, multimodal understanding and captures intricate relationships across modalities, ultimately improving classification performance on complex video tasks.

\begin{figure*}[tbhp]
    \centering
	\includegraphics[width=0.65\linewidth]{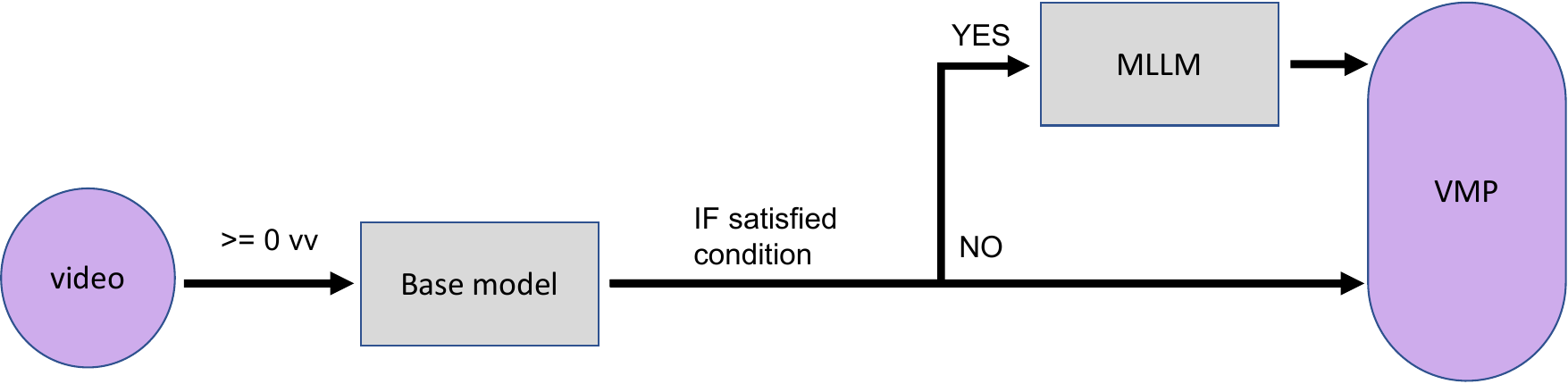}
	\caption{Illustration of Cascade Serving Structure: For the 1st stage, a small, resource-efficient model is used to make initial predictions as soon as a video is posted; for the 2nd stage, only the videos that passed the filtering criteria are subjected to a more thorough evaluation by the MLLM. }
	\label{fig:cascade}
\end{figure*}

\begin{figure*}[tbhp]
    \centering
	\includegraphics[width=0.6\linewidth]{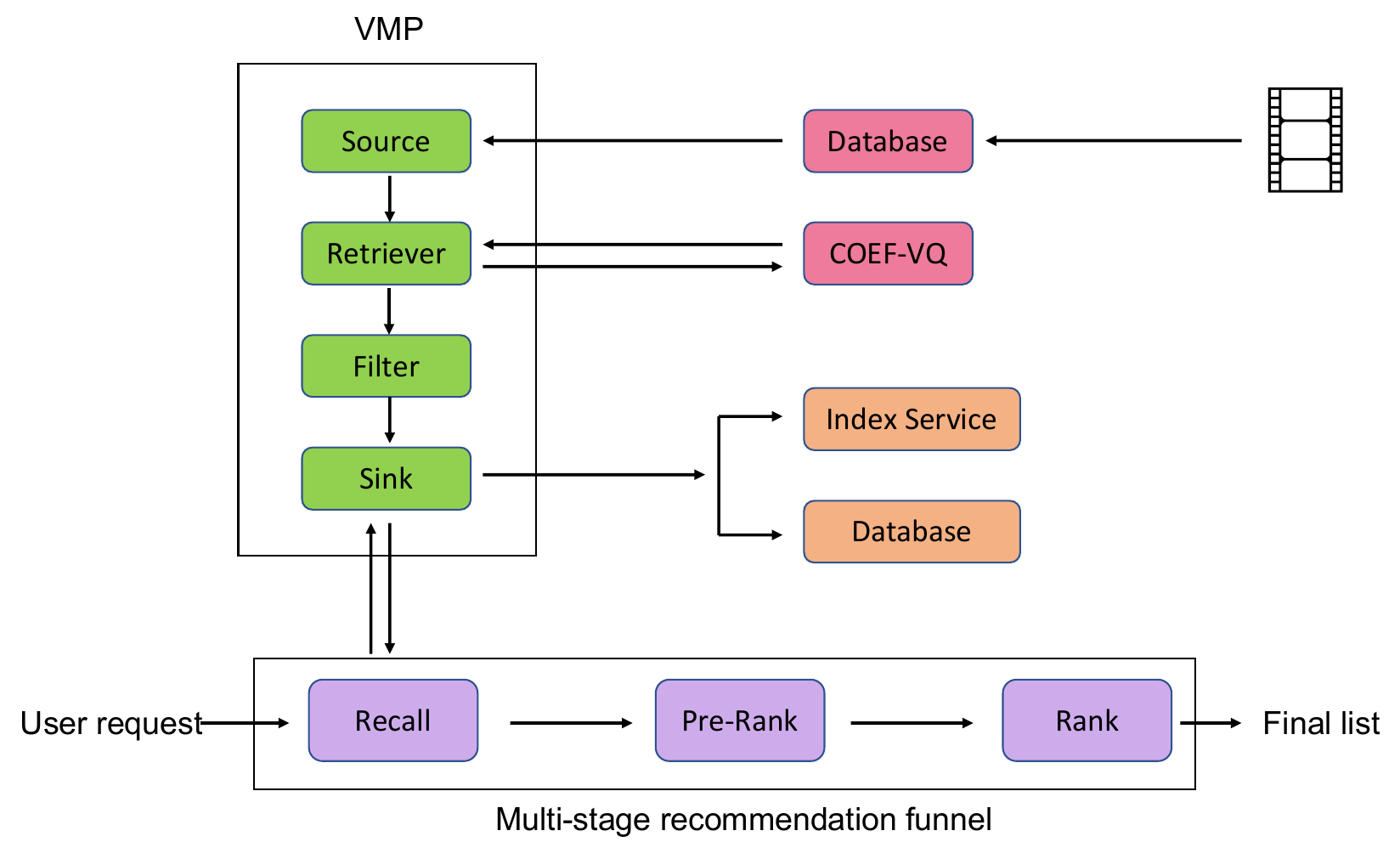}
	\caption{Illustration of Video Management Platform (VMP):  VMP plays a significant role in processing and gatekeeping videos published by creators; the predictions scored in VMP are subsequently processed and utilized in downstream multi-stage recommendation.}
	\label{fig:vcp}
\end{figure*}
\subsection{Cascade Serving Structure}

This framework, as illustrated in \autoref{fig:cascade}, divides the inference process into two distinct stages, each with different computational requirements:

\subsubsection{First Stage: Lightweight Model for Initial Filtering}

In the first stage, a smaller, resource-efficient model is used to make initial predictions as soon as a video is posted. This lightweight model, often designed with fewer parameters and optimized for speed, performs a quick assessment of the content. Here, it assigns a preliminary classification score based on the likelihood of the video containing domain-specific content (such as potentially harmful content). To prioritize informative and ambiguous samples, we propose an entropy-based filtering criterion in our cascade structure. We compute the entropy score \( \textbf{H} \), defined as:  
   \begin{equation}
            \textbf{H}(p) = - \sum_{i=1}^{N} p_i \log p_i
   \end{equation}
where \( p_i \) represents the predicted probability for class \( i \) in an \( N \)-class classification task. 
Videos with low entropy (i.e., confident predictions) are filtered out efficiently, while those with high entropy indicating uncertain or ambiguous cases, are sent to the next stage for further analysis. Additionally, metadata such as video views (VV) or user engagement metrics can be incorporated to further refine the filtering criteria. This stage serves as a gatekeeper, quickly filtering out a significant number of videos with minimal resource consumption, and reducing the workload on the subsequent, more intensive stage.

\subsubsection{Second Stage: Comprehensive Consideration with MLLM}
   In the second stage, only the videos that passed the initial filtering criteria are subjected to a more thorough evaluation by the MLLM. 
   Here, the MLLM re-processes these selected videos, leveraging its full multimodal capabilities to achieve highly accurate classifications. 
   By restricting MLLM’s usage to a narrower subset of content, the two-stage process reduces the heavy usage of GPU to a significant extent if compared with predicting for all contents, 
   while maintaining the high accuracy needed in specialized domains. 
   Moreover, by using the first stage as a triage mechanism, this method allows the MLLM to focus on higher-risk or more ambiguous content, ensuring that its powerful but resource-intensive computations are used judiciously.

\subsubsection{Benefits and Implementation Considerations}
   This cascade deployment structure significantly lowers the GPU resource demands for online applications, making it possible to maintain scalability without incurring prohibitive costs. By using entropy for effective pre-screening, the system reduces the number of inputs needing full MLLM processing, allowing for both cost efficiency and faster response times in high-traffic environments. Furthermore, this method ensures a balance between speed and accuracy, as the MLLM only engages when necessary. In terms of implementation, this cascade structure can be dynamically tuned; thresholds of entropy score in the first stage can be adjusted recurrently based on model performance, further optimizing resource allocation.

\subsection{Deploy \textit{COEF-VQ} onto Video Management Platform (VMP)}
To leverage this newly proposed framework to keep users from inappropriate videos, we deploy \textit{COEF-VQ} onto the Video Management Platform (VMP), such that newly published videos or updated contents could be interacted and safe-guarded through our cascade framework in real-time. 

As a core component of Video Recommendation System, VMP plays a significant role in processing and gatekeeping videos published by creators, to ensure that the most engaging and appropriate content is presented to users.
The functions and key components of VMP are illustrated in \autoref{fig:vcp}.
When a new video is published, model scores related to various quality and policy compliance factors, such as detecting inappropriate content on platforms, are stored in the VMP stage. 
These scores are subsequently processed and utilized in downstream multi-stage recommendation funnel (like Recall, Pre-rank, Rank, etc.).


The VMP operates as a mini-batch streaming framework specifically designed to facilitate the efficient and timely updating of video candidates. The VMP workflow consists of four sequential stages—Source, Retriever, Filter, and Sink—that collaboratively ensure that only high-quality, policy-compliant video candidates proceed to the recommendation stage.

\pheadWithSpace{Source} The Source stage serves as the entry point where video data and stable features are collected and stored. Base data for each video, along with relatively stable features (e.g., metadata, upload date, initial quality assessments), are saved in a Database Server. Upon receiving new videos, the Source stage first generates video IDs by processing KAFKA messages or scanning other data sources. This ensures that each video is uniquely identifiable and can be efficiently processed in later stages.

\pheadNoSpace{Retriever} In the Retriever stage, essential features and metadata are pulled from a variety of services and databases, including Thrift services, Redis caches, and traditional databases. Here, our Cascaded Multimodal LLM Framework is utilized to supply model scores, which represent the video's appropriateness and engagement potential. By incorporating these scores as features, the system gains a nuanced understanding of each video’s quality and relevance. The Retriever thus ensures that all necessary attributes, including multimodal model outputs and contextual metadata, are collected before further processing.

\pheadNoSpace{Filter} During the Filter stage, video candidates are rigorously evaluated based on predefined criteria and divided into three categories: remained, removed, ignored.
This filtering process ensures that only compliant and high-quality content progresses, thereby maintaining a positive user experience. The Filter stage leverages model scores from the Retriever stage to make informed filtering decisions.

\pheadNoSpace{Sink}The Sink stage is responsible for the final updates, storing or removing videos in the Index Service server. In this stage, videos that pass the Filter phase are saved as key-value pairs in the Index Service, where the video ID serves as the key, and all collected features (from model scores to metadata) serve as the value. The Index Service then holds the refined list of video candidates that have been vetted for policy compliance and quality. This final dataset is the primary output of the VMP framework.

By utilizing model scores and other features stored in the Index Service, the recommendation system is empowered to deploy various strategies that align with user preferences and platform policies. 
With the data processed through VMP, the system can dynamically adjust its recommendations, ensuring that the content presented is both engaging and compliant, thus enhancing the overall user experience.

\section{Experiments and Results}

\begin{table*}[tbh]
\centering
\caption{Performance (in \%) from baselines and MLLM on ICD and UCC experiments. MMM: Multi-tower Multimodal Model. The maximum BETA variance across all models for ICD is under 0.08 (in \%), while for UCC is under 0.01 (in \%).}
\label{tab:main-results}
\begin{tabular}{l | c |  c | c c  c  c  c c}
\hline
Task & Approach & Parameter Size & F1&     R@P50 &     R@P60      &     R@P70    &    R@P80 & R@P90 \\ 
\hline
\multirow{2}{*}{ICD}&~~ ResNet50\cite{he2016deep} & 25M & 59.1 & 65.1 & 57.6 & 44.1 & 27.9 &  20.2 \\
&~~ SIGLIP\cite{zhai2023sigmoid}  & 679M & 64.5 & 75.6 & 68.7   & 58.3&  42.9&  29.3 \\
&~~ Qwen2.5\_VL\cite{bai2025qwen2}  & 8.3B & 66.5 & 80.3 & 72.3   & 60.2&  53.7&  39.8 \\
&~~ MLLM (w/o audio)  & 8.3B & 67.8  & 81.1 & 72.4   & 62.2&  53.1 &  39.4 \\
\hline
\multirow{2}{*}{UCC}&~~ MMM & 564M & 77.6& 93.2 &  89.8 &84.5& 74.8 & 54.1 \\
&~~ SIGLIP\cite{zhai2023sigmoid}  & 699M & 75.8 & 91.3 & 87.4   & 80.1&  71.1 &  51.2 \\
&~~ Qwen2.5\_VL\cite{bai2025qwen2}  & 8.4B & 79.5 & 93.4 & 91.9   & 86.9&  78.9 &  61.7 \\
&~~  MLLM & 8.4B & 80.5 & 95.8 & 92.3  & 87.8  & 79.3  & 62.2\\
\hline
\end{tabular}
\end{table*}

\begin{table*}[tbh]
\centering
\caption{Performance (in \%) of \textit{COEF-VQ} on ICD and UCC experiments. QPS Ratio: query per second ratio required for calling MLLM service when video is published in the short-video platform. H\_base: entropy score threshold from the base model in first stage used for filtering.}
\label{tab:cascade}
\begin{tabular}{l | c | c | c  c  c  c}
\hline
Task & $H_{base}$ & QPS Ratio    &  R@P50      &     R@P60    &    R@P70  &    R@P80 \\ 
\hline
\multirow{4}{*}{ICD} & - & 100 & 81.1 &72.4   & 62.2&  53.1  \\
                     & 0.3 & 5.19 & 81.5  & 73.4 &62.6 &52.0 \\
                     & 0.6 & 2.22  & 81.7 & 75.0   & 65.2& 43.9 \\
                     & 0.9 &  0.79  & 79.1 & 71.9   & 58.3& 43.1 \\
\hline
\multirow{4}{*}{UCC}& - & 100 & 95.8 & 92.3  & 87.8  & 79.3  \\
                     & 0.3 &  16.7 & 94.6 & 92.2& 87.3 &79.3 \\
                     & 0.6 & 10.0 & 93.5 & 91.1& 86.7 &79.3\\
                     & 0.9 & 4.55 & 93.0 & 89.9  & 85.8 & 76.5 \\
\hline
\end{tabular}
\end{table*}

\subsection{Experimental settings}
\subsubsection{Datasets} We pick two in-house video quality understanding tasks to test the effectiveness of our proposed MLLM and cascade structure.
The two tasks are Inappropriate Content Detection(ICD) and Unoriginal Content Classification(UCC), respectively.
For ICD, it is a dataset aiming to determine whether the video violates the short-video platform's TnS policy or not, with 1 million training and validation videos and 100K test videos. 
For UCC, it aims to classify whether a posted video is UC(Unoriginal Content) or OC(Original Content). We use a training and validation dataset of 400K videos sampled from the short-video platform and 18K videos as the test set. 
We SFT our MLLM based on each task's training set and tune the hyper-parameters based on each task's validation set. 

\subsubsection{Model Details} 
For MLLM, we invoke the LLaVA-OneVision with 7B model~\cite{li2024llava} as the VL backbone, and consider Whisper-Small\cite{radford2023robust} as the audio encoder. 
We use ZERO2\cite{rajbhandari2020zero} and LoRA\cite{hu2021lora} to train our model. The rank and alpha used in LoRA is 32 and 64. 
For max frame length used in training, we respectively use 8 and 16 frames in ICD and UCC. 
We also rely on left padding when dealing with tokenization. 
For cascade structure, we use the best baseline models as the base model, which are SIGLIP\_Large\cite{zhai2023sigmoid} for ICD task, and a multi-tower multimodal model for UCC task including Swin-transformer\cite{liu2021swin} pretrained by in-house data, XLM-R\cite{conneau2019unsupervised} and Whisper-Base\cite{radford2023robust}. 
\subsubsection{Offline Evaluation Metrics}
Both video quality tasks are evaluated in terms of classification F1 scores, recall at various precision and beta variance.

\textbf{Recall at various Precision} corresponds to the recall rate for binary classification of whether the label is "inappropriate" for ICD task, or "unoriginal" for UCC task.

\textbf{Beta Variance} is defined as the largest variance of a Beta distribution used to model the posterior uncertainty of the positive-class proportion~\cite{NEURIPS2018_a981f2b7}.
Concretely, if a subset of samples has \(\alpha\) for true positives amount + 1 and \(\beta\) for false positives amount + 1, we define variance of a Beta distribution as
\begin{equation}
  \text{Variance} = \frac{\alpha \beta}{(\alpha + \beta)^2 (\alpha + \beta + 1)}
\end{equation}
Then compute this variance for each precision/recall subset and take the maximum as the \emph{maximum Beta variance, reflecting the highest level of uncertainty in estimating the true positive rate}. We consider that maximum Beta Variance below 1\% suggests strong certainty.

\subsection{Main Results}
\pheadNoSpace{Comparisons between baselines and MLLM} 
In Table \ref{tab:main-results}, we present a detailed comparison between MLLM and our baseline models on two classification tasks, demonstrating MLLM’s effectiveness. 
Notably, ResNet50\cite{he2016deep} used for Inappropriate Content Detection (ICD) task operates on a single modality (visual features only); it's a relatively low baseline and so we disable audio encoder in MLLM for ICD task for fairer comparison. 
Besides ResNet50, we also include SIGLIP\_Large\cite{zhai2023sigmoid} and Qwen2.5\_VL\cite{bai2025qwen2} for both tasks as baselines of industry-as-usual. 
However, due to SIGLIP not including pretrained in-house video knowledge, its performance on UCC task is even below our internal multi-tower multimodal model for UCC.

In contrast, MLLM leverages a multimodal architecture that combines visual, audio, and text inputs, enabling it to capture richer, contextualized features that significantly enhance classification accuracy. 
This multimodal integration allows MLLM to achieve a more robust F1 score, capturing nuances that the baseline model might overlook.
Additionally, MLLM shows considerable improvements in recall across a spectrum of precision thresholds, which is especially valuable for content moderation tasks where high precision is critical to minimize false negatives—cases where inappropriate content might go undetected. 
By achieving higher recall while maintaining these precision levels, MLLM reduces the likelihood of such false negatives, ensuring a more reliable detection process in high-precision environments. 

\pheadNoSpace{Results from \textit{COEF-VQ}, the cascade structure} 
Table \ref{tab:cascade} provides a detailed analysis of our cascade structure’s performance across two classification tasks, highlighting its efficiency and precision. 
The cascade framework is designed to selectively trigger the MLLM stage based on predefined entropy thresholds from the base model. 
This selective engagement optimizes resource usage by activating MLLM only when specific entropy criteria are met, allowing the base model to handle simpler cases and reserving MLLM’s capabilities for more ambiguous ones. 

For tasks like Inappropriate Content Detection (ICD), where high recall is essential but often leads to lower precision (due to a greater risk of false positives), the cascade framework plays a critical role. 
By focusing MLLM's processing power on reducing these false positives, the cascade system maximizes detection accuracy while controlling computational costs. 
Specifically, for ICD task, we observe that above an entropy threshold of 0.6 for the base model, the cascade structure actually improves overall performance, achieving results even slightly superior to using MLLM exclusively with a Precision 70 threshold. This finding is notable because it indicates that our cascade approach not only conserves resources but also enhances performance by filtering out false positives with only 2.22\% of the QPS directed towards MLLM. 
Furthermore, for the UCC task, the cascade system achieves performance nearly on par with MLLM alone with 16.7\% of the total query per second (QPS). 
This substantial efficiency gain is achieved without sacrificing performance, demonstrating the cascade’s cost-effectiveness at selectively deploying MLLM only when needed.

The cascade framework significantly reduces computational demands while maintaining accuracy in challenging high-recall, low-precision scenarios. By optimizing the interplay between the base model and MLLM, this approach provides an efficient and scalable solution that adapts well to variable workload demands.

\begin{table}[tbh]
\centering
\caption{Performance (in \%) between \textit{Early fusion} and \textit{Late fusion} for audio modality.}
\label{tab:SFT}
\begin{tabular}{l | c  c  c  c}
\hline
 Approach & F1 &    R@P60    &     R@P70    &    R@P80  \\ 
\hline
~~ \textit{Early fusion}  & 79.9  & 93.2  & 88.2& 78.7  \\
~~ \textit{Late fusion} & 80.5 &92.3 &87.8 &79.3  \\
\hline
\end{tabular}
\end{table}

\begin{table}[tbh]
\centering
\caption{Performance (in \%) from different Rank and Alpha in LoRA.}
\label{tab:lora}
\begin{tabular}{c | c  c  c  c}
\hline
 rank, alpha & F1 &    R@P60    &     R@P70    &    R@P80  \\ 
\hline
0,0 &  65.6 & 69.7 & 60.3 & 48.2  \\
2,4 &  79.7 & 92.4 & 86.9 & 79.5  \\
8,16 &  79.9 &92.8 &87.4 &77.3 \\
32,64 &  80.5 &92.3 &87.8 &79.3  \\
128, 256 &  79.7 &92.6 &86.8 &79.1  \\
512, 1024 &  79.6 & 92.7&88.6 &78.8  \\
\hline
\end{tabular}
\end{table}

\begin{table}[tbh]
\centering
\caption{Performance (in \%) between base model and MLLM with different training sizes on UCC.}
\label{tab:training_data}
\begin{tabular}{c c | c  c  c  c}
\hline
 \#  & Model & F1 &    R@P60    &     R@P70    &    R@P80  \\ 
\hline
20\%  & Base &  73.5 & 85.2 & 77.4 &64.1  \\
   & MLLM & 77.8  & 90.6   & 84.3& 74.4  \\
40\%  & Base &  74.5 & 86.3 & 78.5 & 67.4  \\
   & MLLM & 78.4  & 91.1   & 86.1 & 76.1  \\
60\%  & Base &  76.3 & 87.9 &81.8 & 72.2  \\
   & MLLM & 79.5  & 91.6   & 87.1 & 78.0  \\
100\%  & Base & 77.6&  89.8 &84.5& 74.8 \\
   & MLLM & 80.5  & 92.3  & 87.8  & 79.3 \\
\hline
\end{tabular}
\end{table}

\begin{table}[tbh]
\centering
\caption{Performance (in \%) between \textit{Confidence-based} filtering approach and \textit{Entropy-based} filtering approach with 10\% \textbf{QPS Ratio}.}
\label{tab:entropy}
\begin{tabular}{l | c  c  c  c}
\hline
 Approach &  R@P50      &     R@P60    &    R@P70  &    R@P80  \\ 
\hline
~~ Confidence-based  & 92.9  & 88.4 & 83.1& 77.7  \\
~~ Entropy-based & 93.5 & 91.1& 86.7 &79.3  \\
\hline
\end{tabular}
\end{table}

\subsection{Ablation Studies and Analysis}
In this section, we only consider UCC task for the detailed ablation studies and analysis.

\pheadWithSpace{Comparisons between Audio Fusion Strategies}
We compare the performance of \textit{Late fusion} used in our framework, with \textit{Early fusion}, as shown in Table \ref{tab:SFT}. 
In the \textit{Early fusion} approach, the output from the audio encoder is fed into the LLM as the final input token. 
The results indicate that \textit{Late fusion} achieves a slightly higher F1 score.
Our interpretation is that even though \textit{Early fusion} would provide a stronger feature crossing, it requires a foundation model with better modality alignment between audio and VL, which currently there isn't one.
Therefore, we ended up integrating the audio feature with \textit{Late fusion} due to its proven effectiveness, as well as robustness in terms of audio feature coverage.

\pheadNoSpace{Comparisons between Filtering Criteria} We evaluate the performance of the \textit{Entropy-based} filtering approach in our framework by comparing it with the \textit{Confidence-based} approach, as presented in Table \ref{tab:entropy}. To ensure a fair comparison, we maintain a consistent QPS ratio for both methods. The \textit{Confidence-based} approach selects samples with confidence scores exceeding a predefined threshold.  

The results show that the \textit{Entropy-based} filter achieves higher recall across different precision levels, indicating its effectiveness in selecting more informative cases for the second stage. This demonstrates that our \textit{Entropy-based} method prioritizes uncertain cases more efficiently, leading to improved second stage processing.

\pheadNoSpace{Effects from Diffferent Rank and Alpha in LoRA} To further evaluate the impact of rank and alpha values in LoRA, Table \ref{tab:lora} compares various configurations. We conducted experiments on UCC with progressively increasing rank and alpha values. The setting (32, 64) yielded the best F1 score compared to other configurations. When both rank and alpha are set to 0, indicating that only the classification head is fine-tuned, a poorer result is observed, confirming LoRA’s superiority in enhancing both generalization and robustness.

\pheadNoSpace{Effectiveness on Different Training Sizes} 
MLLM excels in handling situations with limited labeled data. The results in Table \ref{tab:training_data} demonstrate that when fine-tuning MLLM using only part of the UCC training instances, we achieve significantly better performance compared to the base model. 
Additionally, the relative improvements with 20\% of the training data are greater than those observed with 40\% or 60\%, highlighting that MLLM is particularly effective when training data is scarce.

\begin{table}[tbh]
\centering
\caption{Performance (in \%) from ablation studies over MLLM on UCC.}
\label{tab:ablation}
\begin{tabular}{l | c  c  c  c}
\hline
 Approach & F1 &    R@P60    &     R@P70    &    R@P80  \\ 
\hline
~~MLLM & 80.5 &92.3 &87.8 &79.3 \\\hline
~~(a) w/o Whisper &  78.8 &92.3 &86.5 &76.1  \\
~~(b) w/o video text & 78.7  & 91.9   & 84.8 & 77.2  \\
~~(c) w/ Whisper-Base & 79.7  & 92.4   & 88.3 & 77.9  \\
\hline
\end{tabular}
\end{table}

\pheadNoSpace{More ablation studies over MLLM} 
Table \ref{tab:ablation} presents the results from our ablation study on the UCC dataset, examining the impact of different components within our Multimodal Large Language Model (MLLM). In the table, the unmodified MLLM represents our original, fully configured model. The study explores several modifications: (a) removing the Whisper encoder, (b) excluding video text elements (title and stickers), and (c) replacing the Whisper-small audio model with Whisper-base, a smaller parameter model. Each modification isolates specific aspects of the MLLM to assess their relative contributions to classification performance.

In configuration (a), we removed the Whisper encoder, effectively eliminating audio signal processing within the model. This modification resulted in a significant drop in the F1 score, confirming the critical role of audio information in the UCC task. Audio is particularly beneficial in video understanding, as professionally edited videos often include narration that conveys essential content. By incorporating the Whisper encoder, MLLM is better equipped to capture these audio-based cues, which contribute to more nuanced and accurate classification. The results here emphasize that audio features are not merely supplementary; rather, they are integral to comprehensive video understanding, especially in a multimodal context.

Configuration (b) excludes video text components, specifically the video title and stickers, from the model’s inputs. This adjustment led to a nearly 2\% decrease in the F1 score, indicating that text features derived from these elements play a crucial role in the model’s understanding of video content. Video titles and stickers often provide context, keywords, or additional metadata that reinforce visual information, helping the model interpret scenes more effectively. The observed F1 score drop confirms that without these text inputs, the model’s comprehension of the video content becomes less robust, underscoring the importance of text data in multimodal fusion.

In configuration (c), we replace the Whisper-small encoder with Whisper-base, a variant with fewer parameters. By reducing the model size, we observe a noticeable decrease in classification performance, as evidenced by a lower F1 score. This result illustrates the advantage of a larger audio model in capturing the richness of audio cues present in video content. Larger Whisper models, with more parameters, likely capture finer-grained audio details, enhancing the model’s ability to parse subtle audio nuances and supporting more accurate decision-making in classification tasks.

\section{Online Evaluation}
We introduce our proposed cascade framework to the ICD task aimed at reducing inappropriate content on the short-video platform for our online evaluation. 

\pheadWithSpace{Model Deployment}
The MLLM, fine-tuned during offline training, has been deployed online using A10 GPUs with a 24GB memory capacity. For our online A/B test, thanks to our cascade structure, only up to 5\% of the GPUs used for full-traffic MLLM deployment are required.

\pheadNoSpace{Online A/B experiment results}
We conducted online A/B tests, comparing our cascade serving framework (experiment group) with the online setup serving only the base model (control group). The evaluation metric, the inappropriate content video view rate (\textbf{ic\_vvr}), is calculated daily based on human-labeled samples. When a video is identified as inappropriate, a deboosting strategy is applied in the multi-stage recommendation funnel to reduce its visibility, thereby lowering the \textbf{ic\_vvr} platform-wide. Over a 10-day experiment with 10\% traffic allocated to each group, our cascade framework reduced \textbf{ic\_vvr} by \textbf{9.9\%} (statistically significant) with no impact on core metrics.

\pheadNoSpace{Post-launch evaluation} Following the launch of the experiment group, the cascade serving system has been running for more than a month. Global metric monitoring shows a substantial and meet-expectation reduction in \textbf{ic\_vvr}, confirming the effectiveness of the launch.

\section{Related Work}

\pheadNoSpace{Vision-Language Pretraining} 
Recent advances in vision-language pre-training (VLP) have made remarkable strides in aligning visual and textual concepts within a shared latent space. Early VLP approaches, such as those proposed by VinVL \cite{zhang2021vinvl} and ViLBERT \cite{lu2019vilbert}, utilized multimodal transformer encoders to model interactions between visual and language elements. These methods generally depended on pre-extracted features for both images and text and relied on object detectors to associate specific image regions with related text concepts. This approach proved effective in a variety of applications, including visual question answering (VQA), image captioning, and cross-modal retrieval, where a deep integration of visual and textual information is essential.

A significant shift occurred with the introduction of CLIP \cite{radford2021learning}, which demonstrated the effectiveness of using contrastive learning objectives to train dual-encoder models on large-scale, web-sourced image-text datasets. This method has shown excellent performance in tasks such as cross-modal retrieval and zero-shot classification, where alignment between image and text representations is crucial. Following CLIP, researchers have explored several enhancements by combining dual-encoder architectures with different learning techniques.
ALBEF \cite{li2021align} merges image and text embeddings through a multimodal encoder and applies a masked language modeling loss, improving the model's capacity to align multimodal representations. Florence \cite{yuan2021florence} integrates both contrastive and cross-entropy losses to enhance dual-encoder training, effectively balancing between alignment and classification tasks. DeCLIP \cite{li2021supervision} extends this approach further by incorporating contrastive and SimSiam losses, alongside masked language modeling, to boost scalability over standard CLIP models, demonstrating greater flexibility in handling diverse vision-language tasks.

Additionally, SigLIP \cite{zhai2023sigmoid} introduces a simplified approach by employing a pairwise sigmoid loss for image-text pre-training, providing an alternative to conventional contrastive objectives. BEiT3 \cite{wang2022image} proposes a general-purpose multimodal foundation model that adopts a multi-way transformer architecture, integrating both masked image modeling and masked language modeling objectives to support a broad array of vision-language tasks. 

\pheadNoSpace{Audio Modality}
OpenAI's Whisper model \cite{radford2023robust} has demonstrated exceptional speech recognition capabilities by utilizing large-scale weak supervision for speech-to-text tasks. Its training on diverse languages and audio formats enables strong generalization across various recognition challenges, reflecting the growing trend of leveraging massive datasets to enhance deep learning models in audio processing.  
While video-audio~\cite{liu2023cat,zhu2024meerkat} and text-audio~\cite{radford2023robust} models are increasingly prevalent, pretrained models~\cite{liu2023macaw, sun2025audio} that seamlessly integrate all three modalities—video, text, and audio—remain limited, particularly in the context of social media data.

\pheadNoSpace{Multimodal LLM} 
Recent progress in cutting-edge Multimodal Large Language Models (MLLMs), such as GPT-4V\cite{achiam2023gpt}, GPT-4o\cite{achiam2023gpt}, Gemini\cite{team2023gemini}, and Claude-3.5, highlight their impressive adaptability across diverse vision tasks, including single-image, multi-image, and video analysis. 
Many recent research efforts have focused on developing models that are highly optimized for each specific type of task, but there exists a gap in models that can seamlessly handle all three types of tasks at a state-of-the-art level.

To address this gap, LLaVA-OneVision\cite{li2024llava} has been introduced with the goal of achieving top-tier performance across single-image, multi-image, and video tasks, thereby exhibiting robust generalization and effective transfer of learned features in cross-scenario. 
By leveraging advanced techniques for cross-modal alignment and representation, LLaVA-OneVision enables flexible adaptation and accurate predictions across different visual contexts

Several other versatile open models have also shown promising potential for multi-scenario applications in this space. Video-LLaMA\cite{zhang2023video} proposes a multimodal framework that empowers Large Language Models (LLMs) with the capability of understanding both visual and auditory content in the video. VILA\cite{lin2024vila} examined effective pretraining strategies to adapt LLMs for vision-related tasks. InternLMX-Composer-2.5\cite{zhang2024internlm} offers advanced capabilities for handling long-context input and output, supporting features like ultra-high resolution image comprehension, detailed video analysis, multi-turn multi-image dialogue, webpage creation, and article composition. Meanwhile, Qwen-VL series\cite{wang2024qwen2, bai2025qwen2} introduces the Naive Dynamic Resolution mechanism, enabling the model to dynamically process images of varying resolutions into different quantities of visual tokens. Ovis\cite{lu2024ovis} integrates an additional learnable visual embedding table into the visual encoder's process to structurally align visual and textual embeddings.

\section{Conclusion}
Modern video understanding models benefit significantly from a multimodal approach. 
By leveraging the complementary strengths of these modalities, multi-tower architectures achieve robust and holistic performance across general tasks. 
To better address the nuanced demands of specialized applications, 
a unified Multimodal LLM framework can further improve cross-modal interactions and sensitivity to domain-specific cues, 
such as assessing video quality or policy compliance on short-video platforms, 
surpassing traditional multi-tower limitations.

Given the substantial GPU demands of Multimodal LLMs, especially in real-time scenarios, we propose a cascade structure for efficient deployment. 
This two-stage framework begins with a lightweight model that assigns an entropy-based uncertainty score to each video, filtering out low-uncertainty cases before engaging the MLLM for deeper analysis. By prioritizing high-uncertainty inputs, this method significantly reduces computational overhead while maintaining classification accuracy.
Empirical results show that this cascade approach not only optimizes GPU use but also enhances classification precision for specialized tasks, marking a substantial advancement in scalable multimodal video classification. 

\section{Future Exploration}


Future work could extend \textit{COEF-VQ} into a multi-issue MLLM capable of handling various video quality tasks (e.g., quality classification, content moderation, and genre classification) within a single model. This unified approach would streamline processing, enhance consistency, and reduce the need for multiple task-specific models, optimizing resource and maintenance efforts.  

Another key improvement is enhancing audio modality integration. The current late-fused feature concatenation may not fully align text, audio, and visual cues. Future versions could incorporate modality alignment techniques to better capture interactions between these signals, improving the model’s ability to distinguish critical audio cues (e.g., background noise vs. intentional soundtracks) and leading to more precise video quality assessments.
\balance
\bibliography{sample-base}
\bibliographystyle{ACM-Reference-Format}

\end{document}